\pdfoutput=1  
\documentclass[sigconf]{acmart}

\AtBeginDocument{%
  }

\copyrightyear{2026}
\acmYear{2026}
\setcopyright{cc}
\setcctype{by}
\acmConference[MM '26]{Proceedings of the 34th ACM International Conference on Multimedia}{November 10--14, 2026}{Rio de Janeiro, Brazil}
\acmBooktitle{Proceedings of the 34th ACM International Conference on Multimedia (MM '26), November 10--14, 2026, Rio de Janeiro, Brazil}
\acmDOI{10.1145/3767308.3835512}
\acmISBN{979-8-4007-2213-4/2026/11}
\makeatletter
\AtBeginDocument{\@namedef{r@TotPages}{{10}{10}{}{}{}}}
\makeatother

\usepackage{amsmath}
\usepackage{booktabs}
\usepackage{multirow}
\usepackage[table]{xcolor}
\usepackage{graphicx}
\usepackage{makecell}
\usepackage{framed}
\definecolor{shadecolor}{HTML}{EDF2F7}

\newcommand{\Vbar}{\bar{V}}
\newcommand{\zlearner}{\mathbf{z}}
\newcommand{\zdelta}{\mathbf{z}_\Delta}
\newcommand{\zaux}{\mathbf{z}_{\mathrm{aux}}}

\definecolor{cBlue}{HTML}{2B6CB0}
\definecolor{cLightBlue}{HTML}{EBF4FA}
\definecolor{cGray}{HTML}{4A5568}
\definecolor{cGreen}{HTML}{276749}
\definecolor{cOrange}{HTML}{C05621}
\definecolor{cJsonBg}{HTML}{F7FAFC}
\definecolor{cJsonBorder}{HTML}{A0AEC0}
\definecolor{cAccent}{HTML}{3182CE}
\definecolor{cFrozen}{HTML}{90CDF4}
\definecolor{cFrozenFill}{HTML}{EBF8FF}
\definecolor{cTrain}{HTML}{ED8936}
\definecolor{cTrainFill}{HTML}{FFFAF0}
\definecolor{cVLM}{HTML}{A0AEC0}
\definecolor{cVLMFill}{HTML}{F7FAFC}
\definecolor{cToken}{HTML}{48BB78}
\definecolor{cDiffToken}{HTML}{9F7AEA}
\definecolor{cTransfer}{HTML}{3182CE}

\begin{document}

\title{AIDE: Automated Instruction via Distilled Expertise for Reference-Free Motor Skill Coaching}

\author{Yoshiki Ito}
\orcid{0000-0002-8813-9458}
\affiliation{%
  \department{Research \& Development Group}
  \institution{Hitachi, Ltd.}
  \city{Tokyo}
  \country{Japan}}
\email{yoshiki.ito.xf@hitachi.com}


\begin{abstract}
Generating natural-language coaching feedback on motor skills can accelerate learning, yet expert coaches are scarce and expensive.
Existing reference-based methods require expert demonstrations at both training and inference time, limiting practical deployment.
We propose AIDE (Automated Instruction via Distilled Expertise), a framework that exploits expert references only during training and generates feedback from a learner's pose sequence alone at inference.
A teacher model first learns to generate feedback from paired learner--expert poses via a frozen language model, producing separate learner tokens and difference tokens that encode the learner--expert difference.
A student model then inherits the teacher's encoder and weight initialization, replacing the explicit expert comparison with an auxiliary module that produces complementary tokens from the learner's pose alone.
On the ExpertAF dataset, AIDE outperforms reference-free baselines on most metrics and performs comparably to methods requiring expert demonstrations at both training and inference, with LLM-based evaluation supporting these findings.
\end{abstract}




\maketitle

\section{Introduction}
\label{sec:intro}

\begin{figure}[t]
  \centering
  \vspace{2mm}
  \includegraphics[width=\columnwidth]{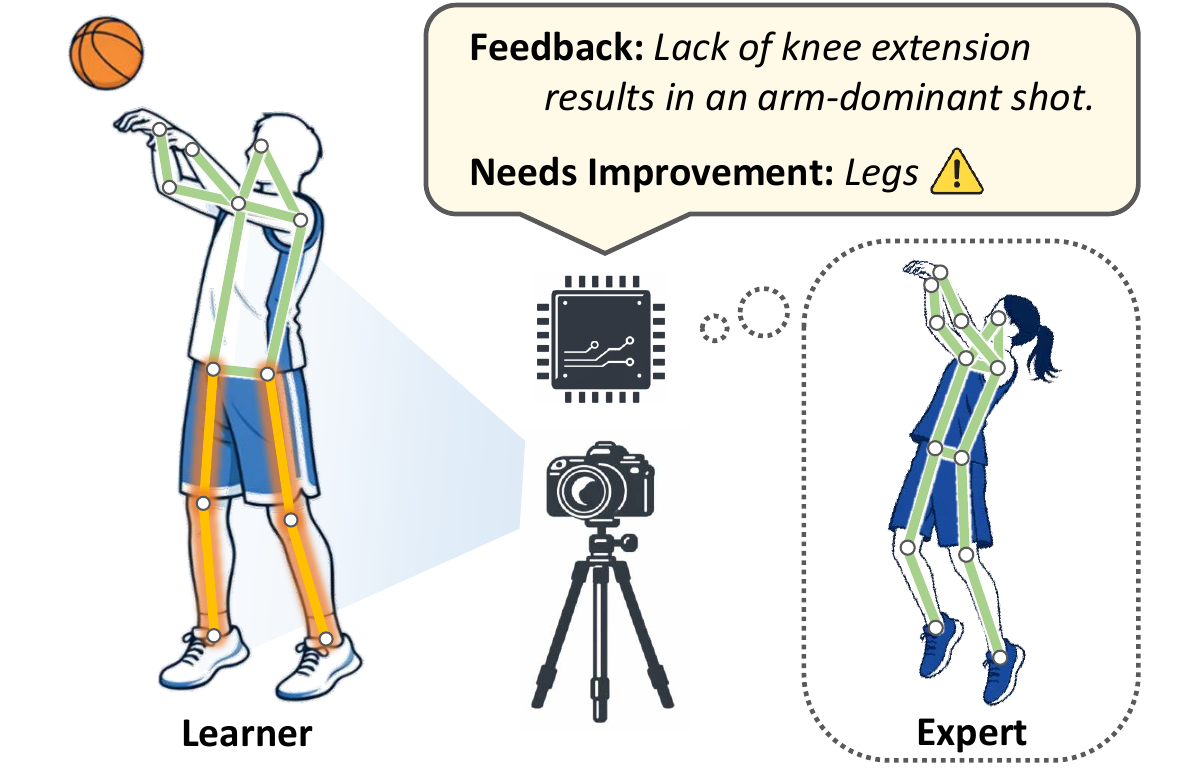}
  \caption{AIDE generates coaching feedback from a learner's pose sequence alone. Expert demonstrations (right) are leveraged only during training; no expert input is required at inference.}
  \vspace{-2mm}
  \label{fig:task_overview}
\end{figure}

Providing specific, actionable feedback on motor skills is essential for learning across domains ranging from youth sports to rehabilitation and industrial training.
However, expert coaches are scarce and expensive: a single coach can observe only a handful of athletes at a time, creating a bottleneck that limits access to quality instruction, particularly in under-resourced communities and remote settings.
Automated feedback generation from video or pose sequences offers a scalable alternative that could make coaching widely accessible.

Recent advances in action quality assessment~\cite{parmar2019aqasurvey, tang2020uncertainty, yu2021group, xu2022finediving} have enabled automatic scoring of athletic performance, and body-pose estimation~\cite{cao2017openpose, sun2019hrnet, xu2022vitpose} now provides reliable skeletal representations for movement analysis.
However, assigning a score is fundamentally different from generating \emph{actionable feedback} that tells a learner \emph{what} to fix and \emph{how}, yet few works tackle this directly.
T3Set~\cite{ma2025t3set} generates feedback for table tennis from learner sensor and pose data, but relies on classifying into predefined suggestion categories rather than producing open-ended text.
ExpertAF~\cite{ashutosh2025expertaf} introduced a paired learner--expert video dataset with structured coaching annotations and a multimodal model that encodes both learner and expert video and pose tokens to generate feedback through a frozen language model.
CoachMe~\cite{yeh2025coachme} generates feedback by concatenating learner and learner--expert difference features from pose sequences, and reports that this expert comparison improves feedback quality over a reference-free variant.
However, these reference-based approaches share a key limitation: they require expert demonstrations at both training \emph{and} inference time, which constrains real-world deployment: obtaining a matched expert demonstration for every learner attempt is impractical.
A retrieval-based alternative, returning a similar learner's stored feedback by nearest neighbor, avoids expert input at inference but reuses annotations and cannot generate learner-specific coaching.
Rather than discarding expert knowledge entirely, we ask whether it can be distilled during training so that inference requires only the learner's input.

We address this limitation through Learning Using Privileged Information~(LUPI)~\cite{vapnik2009lupi, lopezpaz2016unifying}.
LUPI formalizes a setting where additional information (in our case, expert reference poses) is available during training but absent at inference.
This paradigm has been applied to vision tasks~\cite{lambert2018deep, feyereisl2014lupi} but, to our knowledge, not to motor skill coaching.
In our framework, a teacher model first learns from paired learner--expert poses, producing \emph{difference tokens} that encode the learner--expert difference.
A student model then inherits the teacher's encoder and weight initialization, replacing the explicit difference computation with a learned auxiliary module that operates from the learner's pose alone.
We call this framework AIDE (Automated Instruction via Distilled Expertise; Figure~\ref{fig:task_overview}).

A key design choice is the \emph{token-separated architecture}: learner tokens and difference tokens are maintained as distinct sequences throughout the pipeline.
Unlike CoachMe, which concatenates motion and difference features along the feature dimension~\cite{yeh2025coachme}, this separation enables modular knowledge transfer: the student reuses the learner tokens intact while only the difference pathway is retrained.
Our central finding is that this \emph{implicit} knowledge transfer, through shared encoder weights and initialization, without explicit distillation loss, is sufficient to perform comparably to the teacher.
Since the teacher's difference tokens are structurally constrained by the explicit learner--expert subtraction, forcing the student to replicate them would re-impose the limitation it should escape; the student instead learns a representation suited to the end task.

We use a frozen large language model as the backbone, injecting pose-derived tokens via lightweight connectors inspired by the Perceiver Resampler~\cite{alayrac2022flamingo}. Skeleton sequences provide a compact, view-invariant representation, reducing the input to tens of pose-derived tokens and enabling efficient inference without a vision encoder.

We make the following contributions:
\begin{enumerate}
  \item \textbf{LUPI for coaching feedback}: To our knowledge, the first application of LUPI to motor skill coaching. A framework that exploits expert references only during training and generates feedback from the learner's pose alone at inference, eliminating the practical barrier of requiring a matched expert demonstration for every learner attempt.
  \item \textbf{Implicit knowledge transfer}: Encoder sharing and weight initialization alone, without explicit distillation losses, are sufficient to perform comparably to reference-required methods at inference; a capacity control indicates that this is not explained by increased model capacity.
\end{enumerate}

Experiments on ExpertAF~\cite{ashutosh2025expertaf} show that AIDE outperforms reference-free baselines on most metrics on basketball and performs comparably to reference-required methods.
A capacity control experiment indicates that these results are not explained by increased model capacity.
Text quality improvements generalize to soccer, and pairwise LLM-based evaluation on basketball with two independent judges supports these findings.

\begin{figure*}[t]
  \centering
  \includegraphics[width=\textwidth]{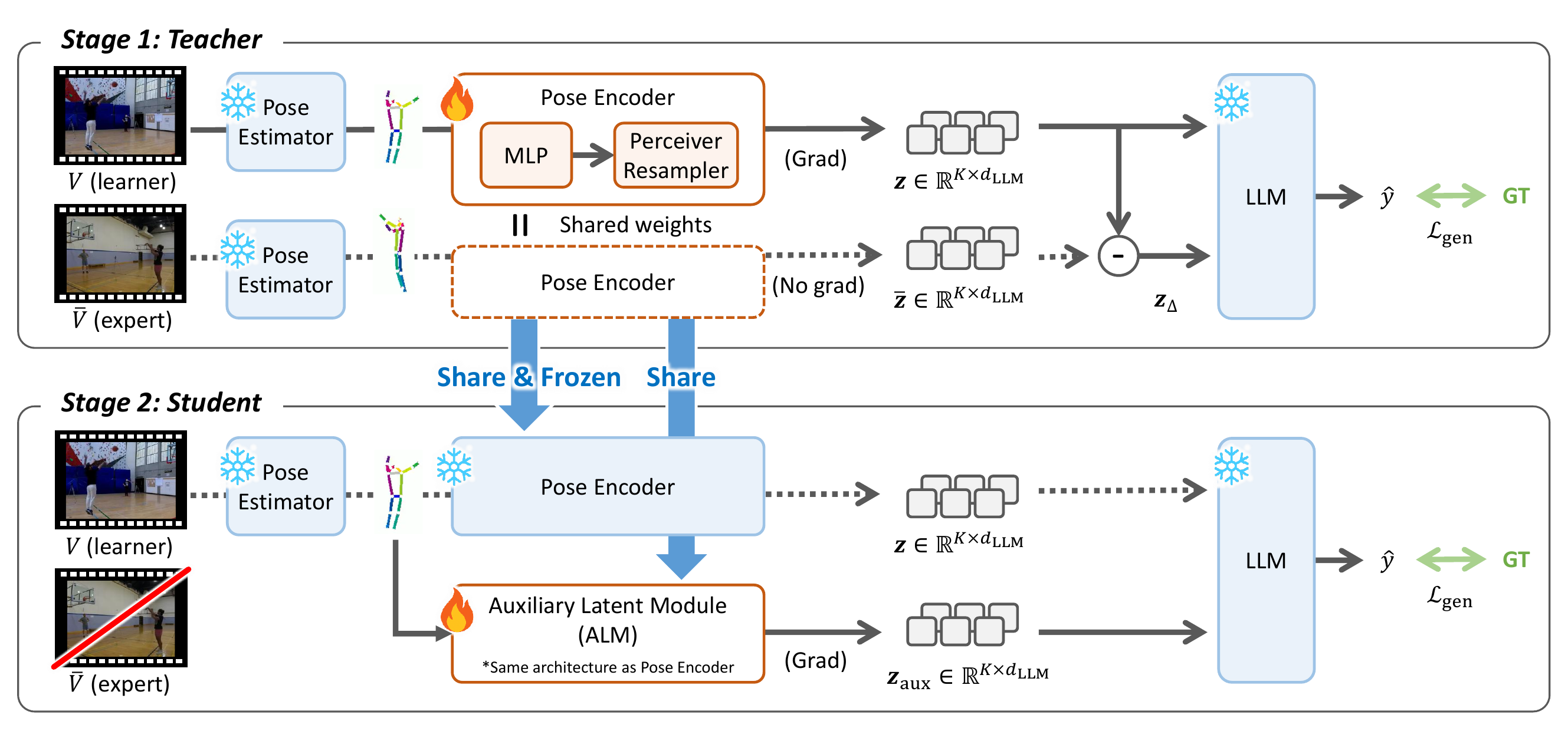}
  \caption{Overview of the AIDE framework. \textbf{Top (Stage~1)}: The teacher encodes both learner $V$ and expert $\Vbar$ through a shared encoder (MLP + Resampler), producing $K$ learner tokens $\zlearner = \mathrm{Enc}(V)$ and $K$ difference tokens $\zdelta = \mathrm{Enc}(V) - \mathrm{Enc}(\Vbar)$. The expert branch is computed without gradients. Both token sequences are concatenated and fed to a frozen LLM. \textbf{Bottom (Stage~2)}: The student reuses the teacher's frozen encoder for $\zlearner$ and a trainable Auxiliary Latent Module (same MLP + Resampler architecture) for $\zaux$, requiring only $V$ at inference. Knowledge transfers via encoder sharing and weight initialization. The trainable module mainly acts as a training-time stabilizer; its inference-time output provides only auxiliary refinement.}
  \label{fig:architecture}
\end{figure*}

\section{Related Work}
\label{sec:related}

\subsection{Coaching Feedback Generation}
Action quality assessment (AQA) methods~\cite{parmar2019aqasurvey, tang2020uncertainty, yu2021group, xu2022finediving} predict numerical scores rather than generating actionable feedback.
ExpertAF~\cite{ashutosh2025expertaf} introduced a large-scale dataset pairing learner--expert sports videos with structured coaching annotations (summary, improvement parts, good parts) and a multimodal model that generates feedback by encoding both learner and expert video and pose tokens through a frozen language model.
CoachMe~\cite{yeh2025coachme} generates feedback by comparing learner and expert poses via a Hierarchical Pose Predictor pre-trained on HumanML3D~\cite{guo2022humanml3d} with T5-base~\cite{raffel2020t5}; it reports that incorporating expert references improves over a reference-free variant, but requires them at both training and inference.
T3Set~\cite{ma2025t3set} provides fine-grained feedback for table tennis from learner sensor and pose data, but focuses on classifying into predefined suggestion categories rather than open-ended text generation.
Other work~\cite{seino2025expertcomment} combines video and motion features with a language model for skill-level-aware comment generation.
Our work bridges reference-based quality and reference-free deployment by exploiting expert poses only during training.

\subsection{Knowledge Distillation and LUPI}
Knowledge distillation (KD)~\cite{hinton2015distilling} transfers knowledge from teacher to student through soft-label matching, feature hints~\cite{romero2015fitnets}, attention transfer~\cite{zagoruyko2017paying}, relational knowledge~\cite{park2019relational}, and contrastive alignment~\cite{tian2020crd}.
Born-Again Networks~\cite{furlanello2018born} showed that students can match or surpass equally-sized teachers, and that explicit feature alignment does not always help~\cite{ojha2023knowledge}, particularly when the teacher's intermediate representations reflect structural constraints that the student should not inherit.
While KD concerns \emph{how} knowledge is transferred between models, Learning Using Privileged Information (LUPI)~\cite{vapnik2009lupi, lopezpaz2016unifying} concerns \emph{what} information is available: it formalizes using additional information only during training, with applications to object localization~\cite{feyereisl2014lupi}, modality transfer~\cite{garcia2018modality}, noisy labels~\cite{collier2022transfer}, and dropout regularization~\cite{lambert2018deep}.
Recent analysis~\cite{rethinking_lupi2024} warns that LUPI gains can be confounded with capacity differences; we address this by using matched architectures across all methods (Section~\ref{sec:experiments}).
To our knowledge, AIDE is the first application of LUPI to motor skill coaching.

\subsection{Multimodal Large Language Models}
Multimodal LLMs (MLLMs) such as LLaVA~\cite{liu2023llava, liu2024llavanext}, BLIP-2~\cite{li2023blip2}, InstructBLIP~\cite{dai2023instructblip}, MiniGPT-4~\cite{zhu2023minigpt4}, and their video extensions~\cite{maaz2024videochatgpt, zhang2023video, lin2024video_llava, qwen2026qwen35, tang2026videounderstanding} have enabled powerful multimodal understanding by bridging non-textual encoders with LLMs.
A key enabler is the use of lightweight connectors that map non-textual representations into the LLM's input space: Flamingo~\cite{alayrac2022flamingo} introduced the Perceiver Resampler for this purpose, LLaVA~\cite{liu2023llava} showed that a simple linear projection suffices for image understanding, and subsequent work~\cite{liu2024llavanext, li2024llava_onevision, lin2024vila} has further scaled these architectures.
Unlike these approaches that process raw visual frames, recent work injects pose or motion features into language models~\cite{yeh2025coachme, seino2025expertcomment, jiang2023motiongpt}.
We follow this direction, injecting pose-derived tokens into a frozen language model via lightweight connectors while training only a small pose encoder.

\section{Method}
\label{sec:method}

\subsection{Overview}

AIDE consists of two stages (Figure~\ref{fig:architecture}).
Both stages use a frozen LLM as the text generation backbone.
Pose-based methods inject tokens directly into the language model, bypassing any vision encoder.
In \emph{Stage~1}, the teacher model is trained on paired learner--expert poses, producing learner tokens $\zlearner$ and difference tokens $\zdelta$ that encode the learner--expert difference.
In \emph{Stage~2}, the student model inherits the teacher's encoder and weight initialization, and learns to generate feedback from the learner's pose alone by replacing the explicit difference computation with a learned Auxiliary Latent Module (ALM).
At inference, only the student model and the learner's pose are required.
This two-stage design follows the Learning Using Privileged Information (LUPI) paradigm: expert references serve as privileged information that improves training but is unavailable at test time.

\subsection{Problem Formulation}

Given a learner's pose sequence $V \in \mathbb{R}^{T \times J \times C}$ ($T$ frames, $J{=}17$ COCO joints, $C{=}6$ channels for position and velocity) and an expert reference $\Vbar$ available only during training,
the goal is to generate coaching feedback text $y$ consisting of a one-sentence summary, body parts needing improvement, and well-executed parts.
During training, both $V$ and $\Vbar$ are available; at inference, only $V$ is given, so the model must internalize comparative knowledge from expert references during training.
This asymmetry distinguishes our setting from standard knowledge distillation, where teacher and student observe the same inputs.

\subsection{Pose Encoder}

Both the teacher and student models use a shared pose encoder architecture $\mathrm{Enc}$ consisting of two components:

\noindent\textbf{Per-frame MLP.}
Each frame's 102-dimensional pose vector ($17 \times 6$) is projected to $d$ dimensions by a two-layer MLP with GELU activation:
\begin{equation}
  h_t = \mathrm{MLP}(v_t) + \mathrm{PE}(t), \quad t = 1, \ldots, T
\end{equation}
where $\mathrm{PE}(t)$ is a sinusoidal positional encoding that provides temporal information.
Following Flamingo~\cite{alayrac2022flamingo}, we process each frame independently and defer temporal aggregation to the Resampler.

\noindent\textbf{Perceiver Resampler.}
The variable-length sequence $H = [h_1, \ldots, h_T]$ is compressed to $K$ fixed-length tokens via a Perceiver Resampler:
\begin{equation}
  \zlearner = \mathrm{CrossAttn}(Q_{\mathrm{latent}},\; H,\; H) \in \mathbb{R}^{K \times d}
\end{equation}
where $Q_{\mathrm{latent}} \in \mathbb{R}^{K \times d}$ are learned latent queries.
The Resampler compresses the variable-length sequence into $K{=}16$ fixed-length tokens, providing a compact interface to the LLM regardless of video duration.
Since pose sequences are far lower-dimensional than visual features, a single-layer Resampler suffices to capture the temporal structure.

An up-projection layer and layer normalization then map the Resampler output from $d$ to $d_{\mathrm{LLM}}$ dimensions.
In subsequent sections, $\zlearner$ denotes the final $d_{\mathrm{LLM}}$-dimensional tokens after this projection.

\subsection{Stage~1: Learner+Expert Teacher}

The teacher model processes both learner and expert poses through the shared encoder:
\begin{align}
  \zlearner &= \mathrm{Enc}(V; \theta_{\mathrm{enc}}) \\
  \zdelta &= \zlearner - \mathrm{sg}\!\left(\mathrm{Enc}(\Vbar; \theta_{\mathrm{enc}})\right) \\
  \hat{y} &= \mathrm{LLM}([\zlearner;\; \zdelta])
\end{align}
where $\zlearner \in \mathbb{R}^{K \times d_{\mathrm{LLM}}}$ are $K$ learner tokens and $\zdelta \in \mathbb{R}^{K \times d_{\mathrm{LLM}}}$ are $K$ difference tokens.
The LLM receives $2K$ tokens in total.
The same encoder (MLP + Resampler) is used for both $V$ and $\Vbar$, ensuring that the subtraction operates in a shared representation space.
We choose element-wise subtraction over cross-attention because it provides an explicit, interpretable difference signal without introducing additional parameters.
The stop-gradient operator $\mathrm{sg}(\cdot)$ detaches the expert branch, so it is treated as a fixed reference in the backward pass: gradients from $\mathcal{L}_{\mathrm{gen}}$ flow back to $\theta_{\mathrm{enc}}$ through both $\zlearner$ and $\zdelta$ via the learner branch, but not through the expert branch.
This prevents the encoder from collapsing to a trivial solution that minimizes $\|\zdelta\|$ rather than capturing the learner--expert gap.

Training uses only the generation loss: $\mathcal{L} = \mathcal{L}_{\mathrm{gen}}(\hat{y}, y)$, where $y$ is the ground-truth feedback; no auxiliary losses are needed since the LLM's generation objective provides sufficient supervision for the encoder.
All LLM parameters are frozen; only the pose encoder parameters $\theta_{\mathrm{enc}}$ are updated.

\noindent\textbf{Token-separated design.}
The learner tokens $\zlearner$ and difference tokens $\zdelta$ are injected into the LLM as separate token sequences via distinct special tokens (\texttt{<|video\_conn|>} and \texttt{<|diff\_conn|>}).
This design differs from CoachMe~\cite{yeh2025coachme}, which concatenates motion and difference features along the feature dimension into a single token sequence.
By maintaining $\zlearner$ and $\zdelta$ as independent token groups, the LLM's self-attention can learn to attend to each role separately.
More importantly, $\zdelta$ becomes the \emph{sole pathway} for expert information, structurally compelling the model to encode the learner--expert gap in these tokens.
This separation is critical for the subsequent knowledge transfer: it allows the student model to inherit $\zlearner$ unchanged while replacing only $\zdelta$ with a learned alternative, which would not be possible if the two signals were entangled in a single representation.

\subsection{Stage~2: Reference-Free Student}

In the second stage, the student model replaces the explicit difference computation with a learned ALM that requires only the learner's pose:
\begin{align}
  \zlearner &= \mathrm{Enc}(V; \theta_{\mathrm{enc}}^{*}) \quad \text{(frozen teacher encoder)} \\
  \zaux &= \mathrm{ALM}(V; \theta_{\mathrm{aux}}) \\
  \hat{y} &= \mathrm{LLM}([\zlearner;\; \zaux])
\end{align}
where $\theta_{\mathrm{enc}}^{*}$ denotes the teacher's trained encoder weights, which are frozen during student training.

\noindent\textbf{Auxiliary Latent Module architecture.}
The ALM uses the same architecture as the pose encoder (per-frame MLP + Perceiver Resampler), ensuring architectural parity across all methods.
The output $\zaux \in \mathbb{R}^{K \times d_{\mathrm{LLM}}}$ occupies the same token slot as the teacher's $\zdelta$, but its content is not constrained to match $\zdelta$.
Rather than reconstructing an explicit learner--expert difference, this module mainly serves as a training-time stabilizer that preserves the teacher's two-pathway interface; its specific inference-time output yields only auxiliary refinements to the generated feedback, as analyzed in Section~\ref{sec:discussion}.

\noindent\textbf{Implicit knowledge transfer.}
The teacher model contributes to the student model through two implicit mechanisms:
\begin{enumerate}
  \item \textbf{Encoder sharing}: The learner tokens $\zlearner$ are produced by the teacher's frozen encoder $\theta_{\mathrm{enc}}^{*}$, which was trained jointly with expert references. This encoder has learned to produce representations that are informative in the context of expert comparison, even when applied to learner poses alone.
  \item \textbf{Weight initialization}: The ALM's weights $\theta_{\mathrm{aux}}$ are initialized from the teacher's encoder $\theta_{\mathrm{enc}}^{*}$, providing a strong starting point that encodes knowledge about the learner--expert relationship.
\end{enumerate}

\noindent\textbf{Training objective.}
Only the generation loss $\mathcal{L} = \mathcal{L}_{\mathrm{gen}}(\hat{y}, y)$ is used.
No explicit distillation losses (e.g., $\|\zaux - \zdelta\|_2^2$) are applied.
Since $\zdelta$ is defined by subtraction from the expert encoding, which is unavailable to the student, exactly reproducing it is neither possible nor desirable.
Instead, the ALM is free to learn any representation that helps the LLM generate accurate feedback, optimizing directly for the end task rather than mimicking the teacher's structurally constrained intermediate representation.
During training, gradients flow through the ALM $\theta_{\mathrm{aux}}$ but not through the frozen encoder $\theta_{\mathrm{enc}}^{*}$.

\subsection{Prompt Design}

Pose-derived tokens are prepended to the text prompt as learner tokens ($K$) and, for dual-pathway methods, difference tokens ($K$).
The LLM is then prompted to generate a structured JSON object.
The AIDE inference prompt is shown below (the full set of prompts for all methods is provided in the supplementary material):

\begin{snugshade}
\noindent\ttfamily
This shows a learner performing \textnormal{\textit{\{sport\}}}.\\
The following tokens encode the learner's body movements extracted from pose estimation.\\
Learner tokens: \textnormal{\textlangle|video\_conn|\textrangle} $\times\;K$\\
The following tokens encode the predicted technique analysis.\\
Analysis tokens: \textnormal{\textlangle|diff\_conn|\textrangle} $\times\;K$\\[4pt]
Output ONLY a valid JSON object with this schema:\\
\{"one\_sentence\_summary": string,\\
\quad "needs\_improvement\_parts": [string, ...],\\
\quad "good\_execution\_parts": [string, ...]\}\\[4pt]
Guidelines:\\
- one\_sentence\_summary: exactly one sentence of expert coaching feedback.\\
- Describe what the athlete is doing and what they should change.\\
- Be specific: refer to concrete body movements.\\
- needs\_improvement\_parts: 0--6 unique items from [Head, Shoulder, Hands, Arms, Legs, Jump].\\
- good\_execution\_parts: 0--6 unique items from [Head, Shoulder, Hands, Arms, Legs, Jump].\\[4pt]
Return JSON only.
\end{snugshade}

\noindent The structured output format enables automatic extraction of body-part predictions for evaluation, while the free-text summary captures the coaching advice.
A key design choice is the distinction between AIDE's ``predicted technique analysis'' tokens and the teacher's ``difference between learner and expert'' tokens: although they occupy the same token slots, the prompt description reflects that the student model generates its analysis independently rather than computing an explicit comparison.

\section{Experiments}
\label{sec:experiments}

\subsection{Dataset}

We evaluate on ExpertAF~\cite{ashutosh2025expertaf}, the only publicly available dataset providing paired learner--expert sports videos with structured coaching feedback.
ExpertAF is built on top of Ego-Exo4D~\cite{grauman2024egoexo4d}, a large-scale multi-view dataset capturing skilled and amateur performances across diverse physical activities.
Each sample contains a learner video, a matched expert reference video, and a coaching annotation derived from expert coaches' spoken commentary: voice recordings are transcribed via ASR and structured into a one-sentence summary, a list of body parts needing improvement, and a list of well-executed parts using Llama~3~\cite{grattafiori2024llama3}. The evaluation set is manually verified.
Body parts are drawn from six categories: Head, Shoulder, Hands, Arms, Legs, and Jump.
Pose sequences are extracted at 17 COCO keypoints with 6 channels per joint (3D position $+$ 3D velocity) at 30\,fps.

Among the sports in ExpertAF, we use the two for which reliable pose sequences are available from Ego-Exo4D:
\begin{itemize}
  \item \textbf{Basketball} (primary): 7{,}476 train / 258 val pairs.
  \item \textbf{Soccer}: 232 train / 71 val pairs.
\end{itemize}

ExpertAF pairs each learner video with 5 different expert references, producing 5 samples that share the same learner video and ground-truth feedback.
These duplicates would inflate evaluation counts; all methods are therefore evaluated on the same deduplicated validation set (basketball: 190; soccer: 62), retaining one sample per unique learner video.
Learner-Only, which discards the expert reference, is trained on the correspondingly deduplicated training set (basketball: 1{,}474; soccer: 59); all other trained models use the full training set.

\subsection{Baselines and Evaluation}
\label{sec:baselines}

We compare against the following methods, all sharing the same frozen LLM backbone (except IV2-NN, which is retrieval-based):

\begin{itemize}
  \item \textbf{Zero-shot}: The VLM processes raw video frames directly without any training, establishing a lower bound.
  \item \textbf{IV2-NN} (nearest-neighbor retrieval): A training-free baseline following the protocol in ExpertAF~\cite{ashutosh2025expertaf}, using Intern\-Video2-Stage2 6B~\cite{wang2024internvideo2} features (16 windows, 768-dim, 224p, 16\,fps). Videos are represented by mean-pooled, centered, $\ell_2$-normalized embeddings; the cosine-nearest training sample's feedback is returned.
  \item \textbf{Learner-Only}: A pose encoder (MLP + Perceiver Resampler) trained on learner poses only, with no expert reference at training or inference. This is the primary baseline for measuring AIDE's improvement.
  \item \textbf{CoachMe}~\cite{yeh2025coachme}: Re-implemented on Qwen3.5 with an MLP-based pose encoder and motion--difference concatenation. The original uses HPP pre-trained on Human\-ML3D~\cite{guo2022humanml3d} with T5-base (223\,M); we replace HPP with a frame-level MLP for fair comparison under a common 9B VLM backbone. Requires reference at both training and inference.
  \item \textbf{Learner+Expert}: AIDE's Stage~1 teacher with explicit difference computation ($\zdelta = \zlearner - \mathrm{Enc}(\Vbar)$). Requires reference at both training and inference.
  \item \textbf{AIDE}: Our proposed framework (Stage~2 student at inference). Reference-free.
\end{itemize}

We evaluate along three complementary axes.
\emph{Text quality} is measured by BLEU~\cite{papineni2002bleu} (B1, B4), ROUGE-L~\cite{lin2004rouge} (R-L), METEOR~\cite{banerjee2005meteor} (M), and BERTScore~\cite{zhang2020bertscore} (BERT; DeBERTa-xlarge-mnli~\cite{he2021deberta}).
\emph{Part prediction accuracy} measures whether the model correctly identifies body parts needing improvement, reported as frequency-weighted F1 (P-wF1), the per-category F1 averaged with weights proportional to each part's support, and exact match (P-EM). P-wF1 is the primary metric as it accounts for the imbalanced part distribution.
Part predictions are extracted by parsing the structured JSON output.
\emph{LLM-based pairwise evaluation} uses two independent judges from different model families (Claude Sonnet 4.6~\cite{anthropic2025claude} and Qwen3.5-9B) to compare AIDE against each baseline on correctness and specificity (190 samples, position-randomized).

\subsection{Implementation Details}

We use Qwen3.5-9B~\cite{qwen2026qwen35} as the base VLM with all parameters frozen; only the pose encoders are trained. We choose a VLM so that the same model can serve as the zero-shot video baseline.
The zero-shot baseline processes raw video frames at 10\,fps (up to 50 frames) with $796 \times 448$ resolution, resulting in ${\sim}7{,}000$ vision tokens per sample.
Training uses AdamW with learning rate $5 \times 10^{-5}$, weight decay $10^{-4}$, gradient clipping at 1.0, linear warmup of 200 steps, dropout 0.1, and pose augmentation.
All connector-based methods share a bottleneck dimension of $d{=}512$, frame MLP hidden dimension of 1024, and $K{=}16$ output tokens. Learner-Only, AIDE, and Learner+Expert use the same Perceiver Resampler (8 attention heads). CoachMe retains its original design with average-pooling instead of the Resampler, with $d$ matched for fair comparison.
All models are trained for 5 epochs with batch size 1 and gradient accumulation of 4.
Inference uses nucleus sampling with temperature 0.5, top-$p$ 0.8, and repetition penalty 1.05.
Results are reported as mean$\pm$std over 3 random seeds.

\begin{table*}[!t]
  \caption{Main comparison on basketball. Train/Infer: whether expert reference is used (\checkmark = used). CoachMe is re-implemented under a common backbone and pose encoder; see Section~\ref{sec:baselines} for details. \textbf{Bold} = best, \underline{underline} = second best.}
  \label{tab:main}
  \setlength{\tabcolsep}{3.5pt}
  \begin{tabular}{l@{\hspace{14pt}}cc@{\hspace{14pt}}ccccccc}
    \toprule
    Method & Train & Infer & BLEU-1 & BLEU-4 & ROUGE-L & METEOR & BERTScore & P-wF1 & P-EM \\
    \midrule
    \rowcolor{black!10} \multicolumn{10}{l}{\emph{No expert reference}} \\
    Zero-shot & & & 22.4 & 0.3 & 14.7 & 14.3 & 75.5 & 30.4 & 6.3 \\
    IV2-NN & & & 24.0 & 2.5 & 19.4 & 18.2 & 75.4 & 39.7 & 7.9 \\
    Learner-Only & & & 26.2\tiny{$\pm$1.0} & 2.7\tiny{$\pm$0.3} & 20.7\tiny{$\pm$0.9} & 19.0\tiny{$\pm$1.4} & 76.1\tiny{$\pm$0.3} & 37.6\tiny{$\pm$5.7} & 12.3\tiny{$\pm$3.0} \\
    \midrule
    \rowcolor{black!10} \multicolumn{10}{l}{\emph{Expert reference at training only (proposed)}} \\
    \textbf{AIDE} & \checkmark & & \textbf{27.9\tiny{$\pm$0.9}} & \textbf{3.9\tiny{$\pm$0.1}} & \textbf{22.3\tiny{$\pm$1.0}} & \underline{20.4\tiny{$\pm$0.5}} & \textbf{76.8\tiny{$\pm$0.6}} & \underline{43.5\tiny{$\pm$5.3}} & \underline{15.8\tiny{$\pm$2.8}} \\
    \midrule
    \rowcolor{black!10} \multicolumn{10}{l}{\emph{Expert reference at training and inference}} \\
    CoachMe & \checkmark & \checkmark & 26.6\tiny{$\pm$2.2} & 2.9\tiny{$\pm$1.4} & \underline{21.8\tiny{$\pm$1.6}} & \textbf{20.6\tiny{$\pm$1.4}} & \underline{76.5\tiny{$\pm$0.5}} & \textbf{45.1\tiny{$\pm$2.4}} & 14.2\tiny{$\pm$0.5} \\
    Learner+Expert & \checkmark & \checkmark & \underline{27.1\tiny{$\pm$0.4}} & \underline{3.4\tiny{$\pm$0.4}} & 21.7\tiny{$\pm$1.2} & 20.1\tiny{$\pm$1.4} & \underline{76.5\tiny{$\pm$0.1}} & 41.8\tiny{$\pm$3.6} & \textbf{16.7\tiny{$\pm$2.7}} \\
    \bottomrule
  \end{tabular}
\end{table*}

\subsection{Main Results}

Table~\ref{tab:main} presents the main comparison on basketball.
Zero-shot VLM performance is low (R-L 14.7), confirming that raw video frames alone are insufficient and that pose-based connectors are essential.
IV2-NN achieves high P-wF1 (39.7) by retrieving the most similar training sample's feedback, but its text quality is limited (R-L 19.4) because it cannot generate novel, sample-specific coaching.

Among trained models, AIDE achieves the highest R-L (22.3) and BERTScore (76.8) despite requiring no reference at inference.
AIDE improves over Learner-Only across all seven metrics (e.g., R-L $+1.6$, P-wF1 $+5.9$) under matched architectural conditions.

Notably, AIDE performs comparably to the Learner+Expert teacher (P-wF1 43.5 vs.\ 41.8, R-L 22.3 vs.\ 21.7) despite having no access to expert references at inference.
The absence of explicit distillation losses (Section~\ref{sec:method}) lets the student optimize directly for feedback quality; in practice, Stage~2's additional training yields an auxiliary improvement in text quality rather than reproducing the teacher's intermediate representation.
CoachMe achieves the highest P-wF1 (45.1) with direct access to expert references at inference; AIDE remains within 1.6 points of it without any expert input, suggesting that inference-time references can be largely replaced by implicit knowledge transfer.

\begin{table}[!t]
  \caption{Results on soccer. Train/Infer: whether expert reference is used (\checkmark = used). \textbf{Bold} = best, \underline{underline} = second best.}
  \label{tab:generalization}
  \setlength{\tabcolsep}{3pt}
  \begin{tabular}{lcccccc}
    \toprule
    Method & Train & Infer & B1 & R-L & BERT & P-wF1 \\
    \midrule
    Learner-Only & & & 21.3\tiny{$\pm$0.5} & 17.6\tiny{$\pm$0.8} & 75.5\tiny{$\pm$0.4} & 46.0\tiny{$\pm$2.5} \\
    \textbf{AIDE} & \checkmark & & \underline{29.6\tiny{$\pm$2.9}} & 26.3\tiny{$\pm$2.7} & \underline{77.7\tiny{$\pm$0.8}} & 38.2\tiny{$\pm$9.8} \\
    CoachMe & \checkmark & \checkmark & 28.7\tiny{$\pm$0.8} & \textbf{28.6\tiny{$\pm$2.3}} & \textbf{78.2\tiny{$\pm$0.6}} & \textbf{49.6\tiny{$\pm$6.3}} \\
    Learner+Expert & \checkmark & \checkmark & \textbf{29.8\tiny{$\pm$4.2}} & \underline{26.6\tiny{$\pm$3.1}} & \underline{77.7\tiny{$\pm$1.1}} & \underline{48.1\tiny{$\pm$7.1}} \\
    \bottomrule
  \end{tabular}
\end{table}

Table~\ref{tab:generalization} evaluates on soccer, a different sport with distinct movement patterns.
AIDE substantially outperforms Learner-Only on all text quality metrics (B1 29.6 vs.\ 21.3, R-L 26.3 vs.\ 17.6, BERT 77.7 vs.\ 75.5), approaching the teacher Learner+Expert (R-L 26.6) without requiring expert references at inference.
P-wF1 is lower for AIDE (38.2) than for all other methods including Learner-Only (46.0), likely due to the small training set of 232 pairs and high cross-seed variance of $\pm$9.8 in this low-data setting, where only 62 validation samples dominated by a single part category make part metrics unreliable; we note this weaker part prediction as a limitation.
The text quality improvements, by contrast, extend to soccer.

\subsection{LLM-Based and Qualitative Evaluation}

\begin{figure}[!t]
  \centering
  \includegraphics[width=\columnwidth]{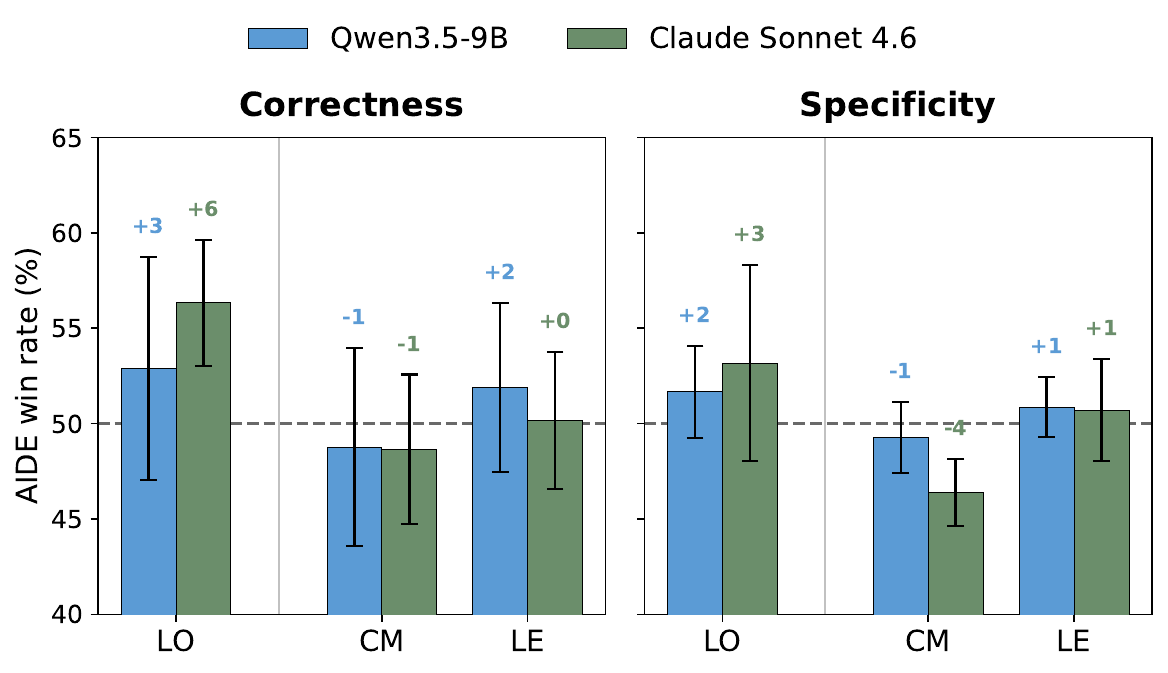}
  \caption{Pairwise LLM-as-a-Judge evaluation: AIDE win rate (\%) against each baseline (LO = Learner-Only, CM = CoachMe, LE = Learner+Expert). Dashed line = 50\% (parity); error bars = std over 3 generation model seeds. Both judges prefer AIDE over LO, while AIDE achieves comparable win rates against CM and LE.}
  \label{fig:geval}
\end{figure}

Following the LLM-as-a-judge framework~\cite{zheng2023judging,liu2023geval}, we conduct pairwise evaluation (Figure~\ref{fig:geval}) using two independent judges from different model families: Claude Sonnet 4.6~\cite{anthropic2025claude} and Qwen3.5-9B.
Since all methods share the same frozen Qwen3.5-9B backbone, any self-preference bias of the Qwen judge affects all methods equally.
To mitigate position bias~\cite{shi2025judging}, the presentation order of candidates is randomized per sample.
Each judge compares AIDE's outputs against each baseline on two independent axes: correctness (are the identified body parts accurate relative to the ground truth?) and specificity (which feedback provides more concrete, actionable coaching advice?).
Both judges prefer AIDE over Learner-Only on both axes, consistent with the automatic metric improvements in Table~\ref{tab:main}.
Against Learner+Expert, both judges produce win rates near or slightly above 50\%; against CoachMe, win rates are near or slightly below 50\%. AIDE achieves comparable feedback quality to methods that use expert references at inference despite not requiring them.

\begin{figure*}[!t]
  \centering
  \includegraphics[width=\textwidth]{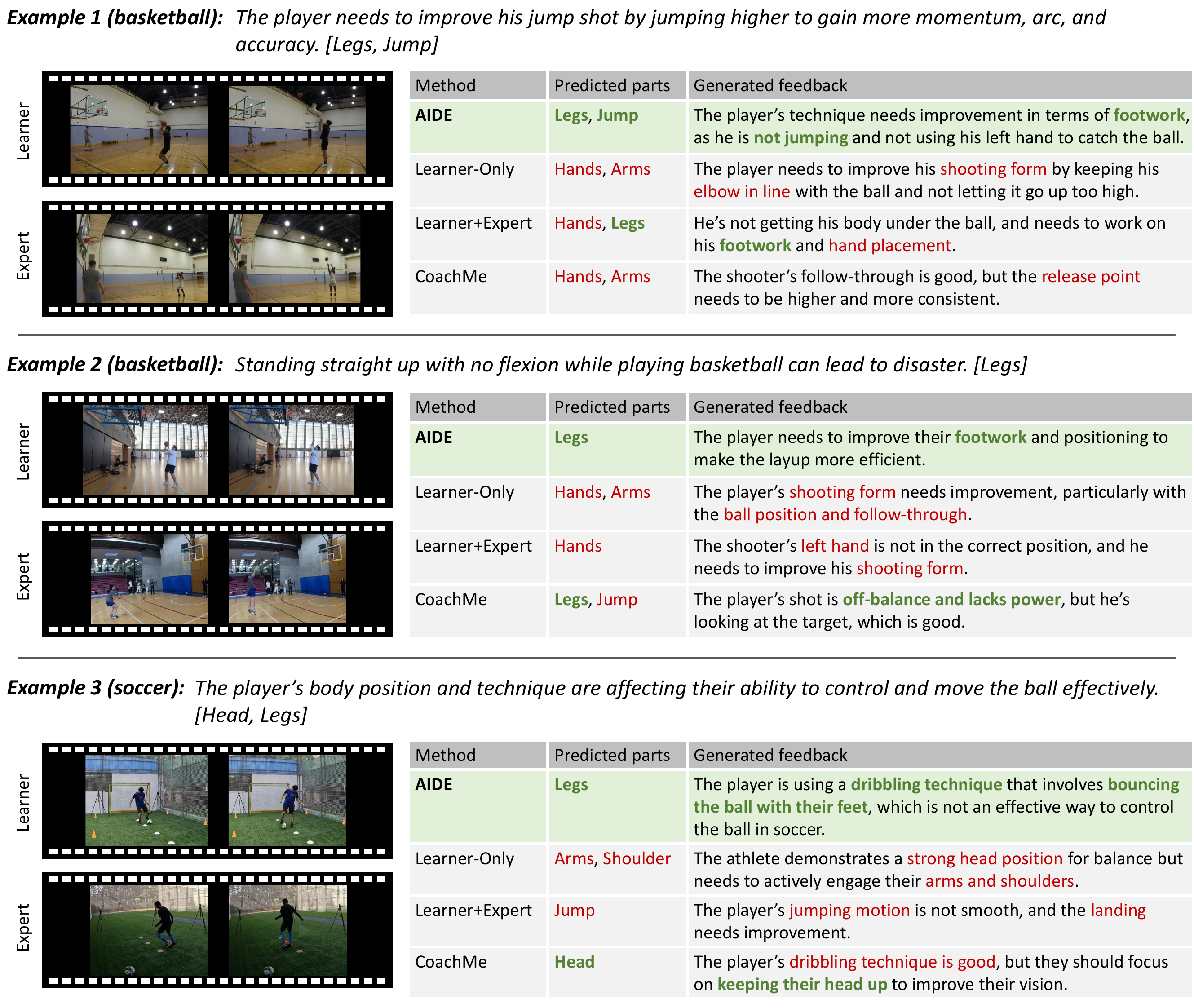}
  \caption{Qualitative comparison on validation samples (2 basketball, 1 soccer). Each example shows the ground-truth feedback with target body parts in brackets. Predicted parts: body parts each method identifies as needing improvement. \textcolor[HTML]{276749}{Green} = correct parts or relevant coaching, \textcolor[HTML]{C53030}{red} = incorrect parts or misleading feedback. Images are cropped for clarity.}
  \label{fig:qualitative}
\end{figure*}

Figure~\ref{fig:qualitative} illustrates representative cases.
In Example~1 (basketball), the ground truth identifies Legs and Jump as needing improvement for the jump shot; AIDE correctly predicts both and notes the player is ``not jumping,'' while Learner-Only and CoachMe predict Hands/Arms, completely missing the lower-body issue.
In Example~2 (basketball), the ground truth identifies Legs (the player is standing straight with no knee flexion); AIDE correctly predicts Legs with relevant coaching about footwork, while Learner-Only predicts Hands/Arms and Learner+Expert predicts Hands, both missing the fundamental stance problem.
Example~3 (soccer) shows that the pattern extends beyond basketball: the ground truth identifies Head and Legs as needing improvement for ball control. AIDE correctly predicts Legs with relevant dribbling advice, though it misses Head (a rare category at 1\% of annotations). Learner-Only predicts Arms/Shoulder and marks Head as ``good execution,'' directly contradicting the ground truth. Learner+Expert describes ``jumping motion'' for a soccer dribbling drill.

\section{Discussion}
\label{sec:discussion}

\subsection{Analysis of Implicit Transfer}
\noindent\textbf{AIDE's gain is not explained by increased capacity.}
Under matched architectures ($d{=}512$ bottleneck for all methods), AIDE outperforms Learner-Only on all seven metrics (Table~\ref{tab:main}).
AIDE inputs 32 tokens (16 learner $+$ 16 difference) vs.\ 16 for Learner-Only, but doubling Learner-Only's tokens to $K{=}32$ (Table~\ref{tab:capacity}) yields only marginal text quality gains (B1 $+$0.4, R-L $+$0.6) and does not improve P-wF1 (33.3 vs.\ 37.6), indicating that AIDE's results are not attributable to increased capacity.

\noindent\textbf{The student model performs comparably to the teacher via implicit transfer.}
AIDE performs comparably to the Learner+Expert teacher (P-wF1 43.5 vs.\ 41.8) despite having no access to expert references at inference, indicating that implicit knowledge transfer through encoder sharing and weight initialization removes the need for inference-time references.
To disentangle these two mechanisms, we train AIDE with a randomly initialized ALM while keeping only the shared encoder (supplementary).
Encoder sharing alone improves text quality over Learner-Only (R-L $+$1.2), but part prediction degrades (P-wF1 36.0 vs.\ 37.6): the randomly initialized ALM produces uninformative tokens that dilute the useful learner signal.
Adding teacher initialization recovers both text quality and part prediction (P-wF1 $+$7.5), confirming that the two mechanisms play distinct roles: encoder sharing provides a common representation space, while weight initialization provides a useful starting point for the ALM.

\noindent\textbf{Implicit transfer suffices.}
Unlike standard distillation~\cite{hinton2015distilling, romero2015fitnets} where the teacher is a larger model, our teacher and student share the same architecture; the teacher's sole advantage is access to expert references, which produce $\zdelta$ via a fixed subtraction.
Since the student cannot access the expert encoding at inference, forcing it to approximate $\zdelta$ would constrain it to a representation it cannot faithfully reproduce, rather than allowing it to discover one better suited to the downstream task.
Empirically, adding an explicit feature-matching loss ($\lambda_{\mathrm{feat}}{=}1$) degrades all metrics (supplementary), confirming that implicit transfer is more effective than explicit loss-based distillation.
This echoes prior findings that explicit feature alignment is not always beneficial~\cite{ojha2023knowledge}; here it fails because the teacher's difference tokens encode a learner--expert subtraction that the student, without expert input, cannot reproduce.
Together with the capacity control and initialization analyses, these results indicate that AIDE's improvement is not explained by increased capacity or explicit distillation.

\subsection{Broader Impact}
Pose-based input provides substantial computational savings over raw video: skeleton sequences reduce the VLM's input from ${\sim}7{,}000$ vision tokens to just 16--32 pose-derived tokens (Table~\ref{tab:tokens}).
AIDE goes further by eliminating expert references at inference, requiring only a single recording of the learner's attempt with 3.5\,M trainable parameters at Stage~2.
This has the potential to democratize access to coaching in settings where expert instruction is scarce or unaffordable, from community sports programs in under-resourced regions to factory floor training in manufacturing.
The structured output format (summary, improvement parts, good parts) is designed to be directly actionable for learners without requiring interpretation by a domain expert.

\subsection{Limitations and Future Directions}
As shown in Section~\ref{sec:experiments}, on soccer where training data is limited (232 pairs), AIDE's text quality gains generalize (R-L $+$8.7) but P-wF1 drops below all methods including Learner-Only (38.2 vs.\ 46.0--49.6). The ALM appears to require more training data for stable part predictions.
Additionally, our evaluation is limited to the ExpertAF dataset, the only publicly available benchmark with paired learner--expert coaching feedback; generalization to other domains remains future work.
All results use a single frozen LLM backbone; transfer to other language models is untested.
Our experiments use the dataset-provided poses from Ego-Exo4D; a deployed system would instead rely on an off-the-shelf pose estimator~\cite{xu2022vitpose}, and the effect of estimation noise on feedback quality remains to be evaluated.
Finally, our evaluation relies on automatic text metrics and LLM-based pairwise judgments; while such protocols have been shown to correlate with human ratings~\cite{zheng2023judging,liu2023geval}, a human expert study would provide stronger evidence of practical coaching utility and remains future work.

Future directions include extending AIDE to other motor skill domains (e.g., rehabilitation, industrial training), integrating end-to-end pose estimation for a fully automatic pipeline, incorporating graph-based pose encoders that exploit joint connectivity, and exploring multimodal encoders that jointly process skeleton sequences with video or sensor signals.

\begin{table}[t]
\centering
\caption{Capacity control: matching AIDE's 32 tokens. LO = Learner-Only.}
\label{tab:capacity}
\setlength{\tabcolsep}{4pt}
\begin{tabular}{lccccc}
\toprule
Method & Tokens & B1 & R-L & BERT & P-wF1 \\
\midrule
LO ($K{=}16$) & 16 & 26.2\tiny{$\pm$1.0} & 20.7\tiny{$\pm$0.9} & 76.1\tiny{$\pm$0.3} & 37.6\tiny{$\pm$5.7} \\
LO ($K{=}32$) & 32 & 26.6\tiny{$\pm$0.6} & 21.3\tiny{$\pm$0.8} & 76.4\tiny{$\pm$0.6} & 33.3\tiny{$\pm$4.8} \\
\textbf{AIDE} ($K{=}16$) & 32 & \textbf{27.9\tiny{$\pm$0.9}} & \textbf{22.3\tiny{$\pm$1.0}} & \textbf{76.8\tiny{$\pm$0.6}} & \textbf{43.5\tiny{$\pm$5.3}} \\
\bottomrule
\end{tabular}
\vspace{3mm}
\end{table}

\begin{table}[t]
\centering
\caption{Computational cost per method.}
\label{tab:tokens}
\begin{tabular}{lccc}
\toprule
Method & Input tokens & Params & Speed$^\ddagger$ \\
\midrule
Zero-shot & $\sim$7{,}000 (vision) & \textendash & $\sim$110\,s \\
Learner-Only & 16 (pose) & 3.5M & $\sim$1\,s \\
CoachMe & 16 (pose) & 3.3M & $\sim$1\,s \\
Learner+Expert & 32 (pose) & 5.4M & $\sim$1\,s \\
\textbf{AIDE} & 32 (pose) & 8.9M$^\dagger$ & $\sim$1\,s \\
\bottomrule
\end{tabular}
\begin{flushleft}
\footnotesize $^\dagger$ 5.4M (Stage~1, teacher) $+$ 3.5M (Stage~2, student). At inference, only Stage~2 is used.\\
$^\ddagger$ VLM inference only. Pose-based methods additionally require pose estimation.
\end{flushleft}
\end{table}

\section{Conclusion}
\label{sec:conclusion}

We presented AIDE, a LUPI-based framework that leverages expert references only during training and generates reference-free motor skill feedback at inference.
AIDE outperforms reference-free baselines on most metrics on basketball, and an ablation indicates that both implicit transfer mechanisms, encoder sharing and weight initialization, contribute to these gains.
Text quality improvements generalize to soccer. Pairwise LLM-based evaluation with two independent judges further supports these findings.
Our central finding is that implicit transfer, without explicit distillation losses, enables a reference-free method to perform comparably to reference-required methods, opening a practical path to scalable coaching from a single learner recording where expert access is limited.

\begin{acks}
This work was supported by JST K Program, Japan Grant Number JPMJKP25V1.
\end{acks}

\clearpage
\bibliographystyle{ACM-Reference-Format}
\bibliography{references}

\balance

\clearpage
\nobalance
\appendix

\suppressfloats[t]

\section*{Supplementary Material}

\renewcommand{\thesection}{\Alph{section}}
\renewcommand{\thesubsection}{\Alph{section}.\arabic{subsection}}
\renewcommand{\thetable}{\Alph{section}\arabic{table}}
\renewcommand{\thefigure}{\Alph{section}\arabic{figure}}
\makeatletter
\@addtoreset{table}{section}
\@addtoreset{figure}{section}
\makeatother

\section{Ablation Studies}
\label{sec:supp_ablations}

\subsection{Token Count Ablation}
\label{sec:supp_k_ablation}

\begin{table}[t]
  \caption{Effect of the number of pose tokens $K$ per pathway on AIDE (basketball). $K{=}16$ is used in all other experiments.}
  \label{tab:k_ablation}
  \setlength{\tabcolsep}{3pt}
  \begin{tabular}{ccccccc}
    \toprule
    $K$ & B1 & R-L & M & BERT & P-wF1 & P-EM \\
    \midrule
    8  & 26.6\tiny{$\pm$1.7} & 20.9\tiny{$\pm$0.5} & \textbf{21.0\tiny{$\pm$1.0}} & 75.8\tiny{$\pm$0.3} & 37.8\tiny{$\pm$5.1} & 10.9\tiny{$\pm$2.7} \\
    \textbf{16} & \textbf{27.9\tiny{$\pm$0.9}} & \textbf{22.3\tiny{$\pm$1.0}} & 20.4\tiny{$\pm$0.5} & \textbf{76.8\tiny{$\pm$0.6}} & \textbf{43.5\tiny{$\pm$5.3}} & \textbf{15.8\tiny{$\pm$2.8}} \\
    32 & 27.6\tiny{$\pm$0.9} & 21.6\tiny{$\pm$0.6} & 20.4\tiny{$\pm$1.7} & 76.6\tiny{$\pm$0.3} & 42.0\tiny{$\pm$5.2} & 13.0\tiny{$\pm$4.3} \\
    \bottomrule
  \end{tabular}
\end{table}

Table~\ref{tab:k_ablation} shows the effect of varying the number of pose tokens $K$ per pathway (learner and analysis) on AIDE.
$K{=}8$ degrades most metrics (B1 $-$1.3, R-L $-$1.4, P-wF1 $-$5.7, P-EM $-$4.9), though METEOR slightly improves ($+$0.6), indicating insufficient capacity for consistent performance.
Doubling to $K{=}32$ yields no improvement over $K{=}16$, with slight declines in R-L ($-$0.7) and P-wF1 ($-$1.5), suggesting that additional tokens introduce redundancy without adding useful information.
$K{=}16$ provides the best overall trade-off, achieving the highest scores on five of six metrics.

\subsection{Teacher Initialization Ablation}
\label{sec:supp_init_ablation}

\begin{table}[t]
  \caption{Effect of teacher initialization on AIDE (basketball). LO = Learner-Only. ``No init'' initializes the Auxiliary Latent Module randomly instead of from the teacher's weights, keeping only encoder sharing.}
  \label{tab:init_ablation}
  \resizebox{\columnwidth}{!}{%
  \begin{tabular}{lcccccc}
    \toprule
     & B1 & R-L & M & BERT & P-wF1 & P-EM \\
    \midrule
    LO              & 26.2\tiny{$\pm$1.0} & 20.7\tiny{$\pm$0.9} & 19.0\tiny{$\pm$1.4} & 76.1\tiny{$\pm$0.3} & 37.6\tiny{$\pm$5.7} & 12.3\tiny{$\pm$3.0} \\
    AIDE (no init)  & 26.8\tiny{$\pm$0.7} & 21.9\tiny{$\pm$0.6} & \textbf{21.2\tiny{$\pm$1.2}} & 76.3\tiny{$\pm$0.3} & 36.0\tiny{$\pm$9.1} & 14.0\tiny{$\pm$4.2} \\
    \textbf{AIDE}   & \textbf{27.9\tiny{$\pm$0.9}} & \textbf{22.3\tiny{$\pm$1.0}} & 20.4\tiny{$\pm$0.5} & \textbf{76.8\tiny{$\pm$0.6}} & \textbf{43.5\tiny{$\pm$5.3}} & \textbf{15.8\tiny{$\pm$2.8}} \\
    \bottomrule
  \end{tabular}}
\end{table}

Table~\ref{tab:init_ablation} isolates the contributions of AIDE's two implicit transfer mechanisms: encoder sharing and weight initialization.
``AIDE (no init)'' uses the same architecture but initializes the Auxiliary Latent Module (ALM) randomly, retaining only the shared encoder.
Encoder sharing alone improves text quality over LO (R-L $+$1.2, M $+$2.2), but part prediction degrades (P-wF1 36.0 vs.\ 37.6 for LO): the randomly initialized ALM produces uninformative tokens that dilute the useful learner signal, leaving the model worse off than LO's 16 focused tokens.
Adding teacher initialization recovers both text quality (B1 $+$1.1, R-L $+$0.4) and part prediction (P-wF1 $+$7.5), confirming that the two mechanisms play distinct roles: encoder sharing provides a shared representation space, while weight initialization provides a useful starting point for the ALM.

\subsection{Distillation Loss Ablation}
\label{sec:supp_loss_ablation}

\begin{table}[t]
  \caption{Effect of adding a feature-matching loss to AIDE. Adding $\mathcal{L}_{\mathrm{feat}}$ degrades both text quality and part prediction.}
  \label{tab:loss_ablation}
  \setlength{\tabcolsep}{3.5pt}
  \begin{tabular}{ccccccc}
    \toprule
    $\lambda_{\mathrm{feat}}$ & B1 & R-L & M & BERT & P-wF1 & P-EM \\
    \midrule
    0 & \textbf{27.9\tiny{$\pm$0.9}} & \textbf{22.3\tiny{$\pm$1.0}} & \textbf{20.4\tiny{$\pm$0.5}} & \textbf{76.8\tiny{$\pm$0.6}} & \textbf{43.5\tiny{$\pm$5.3}} & \textbf{15.8\tiny{$\pm$2.8}} \\
    1 & 27.0\tiny{$\pm$2.1} & 21.6\tiny{$\pm$1.0} & 19.8\tiny{$\pm$1.6} & 76.6\tiny{$\pm$0.1} & 42.3\tiny{$\pm$2.1} & 15.3\tiny{$\pm$2.4} \\
    \bottomrule
  \end{tabular}
\end{table}

Table~\ref{tab:loss_ablation} shows the effect of adding an explicit feature-matching loss ($\mathcal{L}_{\mathrm{feat}} = \|\zaux - \zdelta\|_2^2$) to AIDE's generation loss.
Adding $\mathcal{L}_{\mathrm{feat}}$ ($\lambda_{\mathrm{feat}}{=}1$) degrades text quality (R-L $-$0.7, M $-$0.6) and part prediction (P-wF1 $-$1.2).
Since the teacher's $\zdelta$ is defined by subtraction from the expert encoding, which is unavailable to the student at inference, exactly reproducing it is neither possible nor desirable; the feature-matching loss constrains the ALM toward a structurally limited target rather than allowing it to discover a representation better suited to the generation task.
This confirms that implicit transfer (encoder sharing and weight initialization) is more effective than explicit loss-based distillation.

\section{Prompt Templates}
\label{sec:supp_prompts}

All methods share the same output schema and guidelines (shared suffix below).
We show the method-specific prefix for each variant; $K{=}16$ in all prompts.

\subsection{Shared Output Schema and Guidelines}

The following suffix is appended to all prompts:

\noindent\textbf{Shared Suffix}\\[2pt]
\begin{snugshade}
\noindent\ttfamily
Output ONLY a valid JSON object with this schema:\\
\{"one\_sentence\_summary": string,\\
\quad "needs\_improvement\_parts": [string, ...],\\
\quad "good\_execution\_parts": [string, ...]\}\\[4pt]
Guidelines:\\
- one\_sentence\_summary: exactly one sentence of expert coaching feedback.\\
- Describe what the athlete is doing and what they should change.\\
- Be specific: refer to concrete body movements.\\
- needs\_improvement\_parts: 0--6 unique items from [Head, Shoulder, Hands, Arms, Legs, Jump].\\
- good\_execution\_parts: 0--6 unique items from [Head, Shoulder, Hands, Arms, Legs, Jump].\\[4pt]
Return JSON only.
\end{snugshade}

\subsection{Zero-shot}

\begin{snugshade}
\noindent\ttfamily
This video has been sampled at 10 frames per second.\\[4pt]
\textit{[shared suffix]}
\end{snugshade}

\subsection{Learner-Only}

\begin{snugshade}
\noindent\ttfamily
This shows a learner performing \textnormal{\textit{\{sport\}}}.\\
The following tokens encode the learner's body movements extracted from pose estimation.\\
Pose tokens: \textnormal{\textlangle|video\_conn|\textrangle} $\times\;K$\\[4pt]
\textit{[shared suffix]}
\end{snugshade}

\subsection{AIDE}

\begin{snugshade}
\noindent\ttfamily
This shows a learner performing \textnormal{\textit{\{sport\}}}.\\
The following tokens encode the learner's body movements extracted from pose estimation.\\
Learner tokens: \textnormal{\textlangle|video\_conn|\textrangle} $\times\;K$\\
The following tokens encode the predicted technique analysis.\\
Analysis tokens: \textnormal{\textlangle|diff\_conn|\textrangle} $\times\;K$\\[4pt]
\textit{[shared suffix]}
\end{snugshade}

\subsection{CoachMe}

\begin{snugshade}
\noindent\ttfamily
One shows a learner. Another shows an expert performing the same skill.\\
The following tokens encode the learner's motion and comparison with the expert's technique.\\
Motion-comparison tokens: \textnormal{\textlangle|diff\_conn|\textrangle} $\times\;K$\\[4pt]
\textit{[shared suffix]}
\end{snugshade}

\subsection{Learner+Expert Teacher}

\begin{snugshade}
\noindent\ttfamily
This shows a learner performing \textnormal{\textit{\{sport\}}}.\\
The following tokens encode the learner's body movements extracted from pose estimation.\\
Learner tokens: \textnormal{\textlangle|video\_conn|\textrangle} $\times\;K$\\
The following tokens encode the difference between the learner and an expert reference.\\
Difference tokens: \textnormal{\textlangle|diff\_conn|\textrangle} $\times\;K$\\[4pt]
\textit{[shared suffix]}
\end{snugshade}

\section{LLM-as-a-Judge Protocol}
\label{sec:supp_geval}

We evaluate feedback quality via pairwise LLM judgments.
Given a ground-truth reference and two candidate feedbacks, the judge evaluates two independent axes (correctness and specificity) and is forced to choose a winner on each axis (no ties).

The system prompt provided to each judge is:

\begin{snugshade}
\noindent\ttfamily
You are an expert sports coach evaluating automated coaching feedback. You will be given a ground-truth reference and two candidate feedbacks (A and B) for the same sports performance. Compare A and B on two INDEPENDENT axes. Judge each axis separately.\\[4pt]
You MUST choose either A or B for each axis. Do NOT output Tie.\\[4pt]
Axis definitions:\\
1. \textbf{Correctness}: Which feedback more accurately identifies the body parts and issues described in the ground-truth? Consider both the predicted body parts (Needs improvement / Good execution) and whether the summary text addresses the same problem as the ground truth.\\
2. \textbf{Specificity}: Which feedback provides more concrete, actionable coaching advice? For example, ``keep your elbow aligned with your shoulder'' is more specific than ``improve your form''. Judge this INDEPENDENTLY of correctness.\\[4pt]
Output valid JSON with keys: ``correctness'', ``specificity'' (each ``A'' or ``B''), and ``reasoning'' (1--2 sentences). Output ONLY the JSON object.
\end{snugshade}

The presentation order of A and B is randomized per sample to mitigate position bias.

\end{document}